\documentclass[11pt]{article}

\usepackage[T1]{fontenc}
\usepackage[utf8]{inputenc}   
\usepackage{textcomp}         

\usepackage[super,sort,comma]{natbib}

\usepackage{placeins}
\usepackage{graphicx}
\usepackage{adjustbox}
\usepackage{caption}
\usepackage{subcaption}
\usepackage{amsmath}
\usepackage{amsfonts}

\makeatletter
\renewcommand\@biblabel[1]{$^{#1}$}
\makeatother
\usepackage[a4paper, total={17cm, 22.6cm}]{geometry}

\newcommand{\cen}[1]{\begin{center} #1 \end{center}}

\usepackage{xcolor}
\definecolor{customGray}{rgb}{0.6,0.6,0.6}
\definecolor{customRed}{rgb}{0.85,0,0}
\definecolor{customGreen}{rgb}{0,0.85,0}
\definecolor{customBlue}{rgb}{0,0,0.85}
\definecolor{customBeige}{rgb}{0.92,0.87,0.78}

\usepackage[mathlines]{lineno}

\usepackage{hyperref}
\hypersetup{
    colorlinks,
    citecolor=blue,
    filecolor=blue,
    linkcolor=blue,
    urlcolor=blue,
    pdftitle={Dual-domain U-Nets with embedded back projection operators for motion-resolved 4D CBCT reconstruction},
    pdfauthor={Ivo Herzig, Pascal Paysan, Daniel Barco, Marc Andr\'e Stadelmann, Frank-Peter Schilling, Igor Peterlik, Michal Walczak, Lijin Aryananda, Woo Sang Ahn, Rudolf Marcel F\"uchslin, Lukas Lichtensteiger},
    pdfsubject={Medical Physics, 4D Cone Beam CT, Deep Learning},
    pdfkeywords={4D CBCT; deep learning; motion modeling; image-guided radiation therapy}
}

\begin{document}
\twocolumn[

\cen{\sf {\Large {\bfseries Dual-domain U-Nets with embedded back projection operators for motion-resolved 4D CBCT reconstruction}}
\vspace*{8mm}

Ivo Herzig\textsuperscript{1},
Pascal Paysan\textsuperscript{3},
Daniel Barco\textsuperscript{2},
Marc Andr\'e Stadelmann\textsuperscript{2},
Frank-Peter Schilling\textsuperscript{2},
Igor Peterlik\textsuperscript{3},
Michal Walczak\textsuperscript{3},
Lijin Aryananda\textsuperscript{1},
Woo Sang Ahn\textsuperscript{4},
Rudolf Marcel F\"uchslin\textsuperscript{1},
Lukas Lichtensteiger\textsuperscript{1}
}

\small
\textsuperscript{1} Zurich University of Applied Sciences ZHAW, Institute for Applied Mathematics and Physics IAMP, Winterthur, Switzerland\\
\textsuperscript{2} Zurich University of Applied Sciences ZHAW, Centre for Artificial Intelligence CAI, Winterthur, Switzerland\\
\textsuperscript{3} Varian Medical Systems Imaging Laboratory GmbH, Baden-D\"attwil, Switzerland\\
\textsuperscript{4} Radiation Oncology, Gangneung Asan Hospital, University of Ulsan College of Medicine, Republic of Korea

\pagenumbering{roman}
\setcounter{page}{1}
\pagestyle{plain}

\begin{abstract}

\noindent {\bf Background: }Four-dimensional cone beam computed tomography (4D CBCT) plays an important role in image-guided radiation therapy (IGRT) of thoracic cancers. However, its clinical utility is hindered by long scan times, resulting in high patient dose and compromised image quality such as streaking artifacts caused by respiratory motion and sparse sampling of projections.

{\bf Purpose: } We propose a novel deep learning-based method to enable motion-resolved 4D CBCT reconstruction from conventional clinical free-breathing scans without the need for a respiratory signal or explicit projection binning. The proposed deep convolutional neural network (CNN) takes conventional free breathing 3D CBCT projection data as input and predicts a static volume representing the maximum inhalation phase and a set of ten displacement vector fields (DVFs) representing patient motion during a complete breathing cycle.

{\bf Methods: }
The network extends the well-known U-Net architecture. The downsampling encoder path acts directly on filtered CBCT projection stacks, whereas the upsampling-decoder part acts in the volume domain. The skip connections found in conventional U-Nets are replaced with non-trainable back projection functions at multiple levels of resolution to effectively transfer features from the projection to the volume domain. The model is trained in a supervised fashion on simulated CBCT scans. The method is evaluated on simulated scans of 11 held-out unseen patients from the same simulation dataset, and on 13 clinical free-breathing CBCT scans. Two additional models with different acquisition settings (scan durations of 60\,s and 6\,s) were trained and evaluated using three and two scans, respectively, in a clinical expert evaluation. The experts were presented with side-by-side views of a single phase from our 4D reconstruction and a reference 3D SART-TV image and indicated their preference with regard to tumor and esophagus visibility.

{\bf Results: }
Clinical experts attested improved tumor visibility (59\%, vs.\ 36\% no preference, 5\% preferred a reference 3D reconstruction), and esophagus (47\%, vs.\ 42\% and 11\%). Using the simulated test dataset, we observed stable image quality metrics compared to a traditional SART-TV 3D reconstruction (mean RMSE: -1.19 HU, PSNR: +0.09 dB, SSIM: -0.009) while enabling motion-resolved 4D reconstruction. For clinical CBCT scans, where ground-truth is unavailable, our method demonstrates improved resolution of dynamic structures (e.g., diaphragm) while greatly mitigating motion streak artifacts when compared to traditional reconstructions.

{\bf Conclusions: }
Our proposed novel, non-patient-specific CNN model enables the prediction of static volumes and complete 4D respiratory motion models from a conventional clinical free-breathing scan without the need for a recorded respiratory surrogate signal or explicit projection binning. Our study serves as a proof-of-concept for this method, showcasing reduced motion artifacts in images while adding motion modeling capability.

\end{abstract}
]


\clearpage
\pagenumbering{arabic}
\setcounter{page}{1}
\section{Introduction}

Four-dimensional cone beam computed tomography (4D CBCT) has great potential in Image-Guided Radiation Therapy (IGRT) for thoracic cancers, enabling motion-resolved tumor localization and adaptive treatment planning.

However, the required long acquisition times can result in higher patient radiation dose, and 4D CBCT image acquisition protocols typically require additional hardware and setup time to allow recording of respiratory surrogate signals, further reducing clinical utility \cite{nardiMotion2016}.

Furthermore, reconstruction algorithms, such as the Feldkamp-Davis-Kress (FDK) method \cite{feldkampPracticalConebeamAlgorithm1984} or algebraic reconstruction techniques (ART) \cite{gordonAlgebraicReconstructionTechniques1970} are sensitive to motion corruption and incomplete data, leading to degraded image quality.

We introduce a novel deep learning (DL) approach for motion-resolved 4D CBCT reconstruction from standard clinical free-breathing scans, eliminating the need for a recorded respiratory signal or explicit projection binning \cite{sonkeRespiratoryCorrelatedCone2005}. A 4D CBCT method, which does not require respiratory signals or explicit binning, can improve robustness, efficiency, and image quality by simplifying workflow and reducing errors from surrogate signals.

\subsection{Related Work}
Numerous approaches have been proposed to address motion artifacts and enable time-resolved (4D) CBCT reconstruction, including post-processing of reconstructed images, deep learning-based deformable image registration, projection-binned reconstructions, dual-domain methods, and reconstruction strategies that avoid explicit projection binning \cite{zhangReview4DConebeam2024, amirianReview_2024}.

Post-processing methods operate on reconstructed 3D or 4D volumes and have been used to suppress various types of image defects. We focus on artifacts introduced by motion or sparse sampling. Typically, these methods work by learning mappings from artifact-corrupted to artifact-free images \cite{kidaConeBeamComputed2018}. Types of architecture include CNNs \cite{amirianMitigationMotioninducedArtifacts2023} and generative methods such as generative adversarial networks (GANs) and diffusion models \cite{qiuDeepLearningbasedThoracic2021, dongBetterConeBeamCT2025, huCBCTCTSynthesisUsing2025}.
While many 3D artifact-reduction methods can also be applied to 4D CBCT, some studies specifically focus on the special properties of 4D scans \cite{madestaSelfcontainedDeepLearningbased2020, dengRSTAR4DRotationalStreak2025}.

Motion compensated (MoCo) approaches start with an initial (usually phase correlated) 4D reconstruction and then use motion modeling to combine the 4D volumes to a new set of volumes with reduced sparse-sampling artifacts \cite{ritFlyMotioncompensatedConebeam2009, ritComparativeStudyRespiratory2011, zhangDeepLearningbasedMotion2023, zhangFastMotioncompensatedReconstruction2024}. Deformable image registration (DIR) is often used for motion modeling. One widely adopted framework is VoxelMorph, which employs a U-Net architecture to perform DIR \cite{balakrishnanVoxelMorphLearningFramework2019}. Extensions to groupwise registration and settings with sparse or artifact-laden data have also been studied \cite{zhangGroupRegNetGroupwiseOneshot2021, herzigDeepLearningbasedSimultaneous2022}.

Most 4D techniques, including MoCo, require the binning of projections according to motion state (most often respiratory phases). This requires a respiratory surrogate signal which is recorded during the scan using, e.g., an optical marker block (such as the Varian Real-Time Position Management\textregistered\ (RPM) system), belts, or other optical methods. However, the need for such external equipment increases scan setup complexity and duration. Alternatively, surrogate signals can also be extracted after the scan from projection data, for example using the well-known Amsterdam Shroud \cite{zijpExtractionRespiratorySignal2004} or intensity analysis method \cite{kavanaghObtainingBreathingPatterns2009}. For these methods, however, the accuracy of the signal depends on the quality and properties of the projection data, and usually only works well for a limited range of scan angles.

Several methods have been proposed to combine operations in both the projection and volume domains \cite{gaoTransformerbasedDualdomainNetwork2023}. One such technique incorporates differentiable non-trainable projection operators directly into neural networks \cite{wurflDeepLearningComputed2018}. Other dual-domain strategies have jointly optimized projection interpolation and image-domain enhancement in a unified network \cite{zhaoDLPVIDeepLearning2025} or integrate deep learning components into iterative reconstruction pipelines \cite{chenAirNetFusedAnalytical2020, chen4DAirNetTemporallyresolvedCBCT2020}.

Methods that do not require a surrogate signal are attractive in clinical settings, where such signals are often unavailable. Recent work has explored reconstruction of dynamic 4D CBCT techniques that aim to reconstruct volumes at each projection, bypassing the need for explicit projection binning \cite{jailinProjectionbasedDynamicTomography2021a, huangSurrogatedrivenRespiratoryMotion2024}. Patient-specific learning-based models have been developed to reconstruct images from single projections by estimating deformation models that map a reference image to the observed motion state using explicit motion fields \cite{zakeri4DPreciseLearningbased3D2024} or an implicit neural fields based approach \cite{shaoDynamicCBCTImaging2024}. 

\begin{figure*}[t]
\begin{center}
\includegraphics[width=1.0\linewidth]{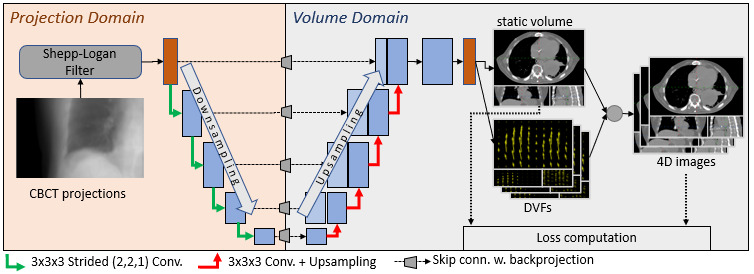}
\end{center}
\caption[architecture]{ \label{fig:architecture}
Overview of the CNN-based reconstruction method. Starting with CBCT projections from a scan and finally predicting a static volume and a set of cyclic DVFs representing patient motion during a breathing period.}
\end{figure*}

\subsection{Contributions}
We propose a deep learning-based method to enable motion-resolved 4D CBCT reconstruction from conventional clinical free-breathing scans without the need for a recorded respiratory surrogate signal and without any explicit projection binning. The neural network is a novel style of dual-domain U-Net with non-trainable back projection operations within its skip connections to allow end-to-end joint motion modeling and 4D image reconstruction directly from projections. The model is not patient-specific and can be trained and evaluated on a mid-tier GPU.

\section{Materials and Methods}

\subsection{Synthetic data generation} \label{sec:synthgen}
Our models are trained on synthetic free-breathing CBCT scans, generated using respiratory motion simulation based on a previously published method \cite{amirianMitigationMotioninducedArtifacts2023}.

DIR is performed across the ten phases of four-dimensional computed tomography (4D CT) using a registration neural network \cite{balakrishnanVoxelMorphLearningFramework2019, herzigDeepLearningbasedSimultaneous2022}. The network registers each phase to the next in a cyclic manner, wrapping from the final phase back to the initial phase, resulting in a series of patient-specific deformation vector fields (DVFs).

Each synthetic scan is generated from a set of DVFs and the corresponding maximum-inhale reference volume combined with a randomly selected respiratory amplitude signal. For each respiratory cycle within this signal, the amplitude is converted into a phase signal and a per-cycle amplitude ratio. The amplitude ratio measures how much a given cycle's inhale-to-exhale amplitude compares to the largest such amplitude in the scan.

During simulated X-ray acquisition, the DVF corresponding to the current phase of the respiratory signal is selected and scaled by the per-cycle amplitude ratio. The reference volume is deformed using this scaled DVF and then forward-projected to generate digitally reconstructed radiographs (DRRs). The result is a set of synthetic CBCT projections that contain realistic respiratory motion patterns with inter-phase variation in amplitude.

Additionally, motion artifact-free ground-truth volumes are generated by deforming the reference volume using the DVFs at ten discrete phases (ranging from 0\% to 90\%). These ground-truth volumes correspond to the breathing cycle with the highest inhale-to-exhale amplitude in the scan.

\subsection{U-Net with embedded back projection operator}
The proposed neural network takes a set of free-breathing CBCT projections as input and predicts a static volume (max-inhale) and a set of circular DVFs that model a complete respiratory cycle (Figure \ref{fig:architecture}). The DVFs are applied to the static volume to create ten breathing phases, resulting in a 4D reconstruction consistent in format with traditional phase-correlated 4D CBCT methods. The network architecture is based on the original U-Net design, adapted to process projection data and output volumetric motion representations. We use six hierarchical levels.

\noindent{\bf Projection pre-processing: }
Following the traditional FDK reconstruction approach \cite{feldkampPracticalConebeamAlgorithm1984}, the scatter-corrected input projections are converted to attenuation space and filtered using a Shepp-Logan filter. The resulting projection stack is standardized to zero mean and unit variance before being fed into the network.

\noindent{\bf Projection domain encoder: }
The projections are arranged into a 3D tensor (sinogram) of dimensions $N \times H \times W$, where $N$ is the number of projections (temporal dimension), and $H,W$ are the spatial projection height and width, respectively. The U-Net encoder uses strided 3D convolutions for hierarchical feature extraction. A stride of 2 is applied along the spatial dimensions at each level to progressively reduce resolution. Along the temporal axis, the first two encoding levels use a stride of 2 to expand the temporal receptive field, while the remaining four levels use a stride of 1 to mitigate sparse-sampling artifacts in the embedded back projection operators. The number of filters at each level is 32, 64, 64, 64, 64, and 64, respectively.

\noindent{\bf Projection-to-volume skip connections: }
At each level of the U-Net, the projection domain feature maps are translated to the volume domain via embedded non-trainable back projection functions \cite{wurflDeepLearningComputed2018}. Instance normalization is applied to the backprojected feature maps and they are then concatenated with the corresponding features in the volume-domain decoder path. The back projection functions are implemented as custom CUDA kernels. When combined with the Shepp-Logan filter applied to the input projections, this resembles the FDK reconstruction method. To allow end-to-end training of the network, the backward pass of these layers is implemented as a forward projection operation.

\noindent{\bf Volume domain decoder: }
Each level applies a block consisting of $3\times3\times3$ convolutions followed by Swish activation \cite{ramachandranSearchingActivationFunctions2017} and 2x trilinear upsampling. In addition to the upsampling layers, we add three additional convolution layers without upsampling on the top level of the network to refine the resulting volumetric outputs at the final output resolution. All convolutional layers in the decoder use 64 filters, making the down- and upsampling paths of the U-Net symmetric except for the top level.

\noindent{\bf Network outputs and volume warping: }
The last layer of the network is a convolutional layer without an activation function; it maps the feature maps to the output volume with dimensions $31\times D\times H\times W$. This output volume is split into a static image ($1\times D \times H \times W$) and a set of 10 DVFs ($10\times D \times H \times W \times 3$), where $D, H, W$ are the volume sizes in $z, y, x$ directions, respectively. Each DVF maps the volume from one breathing phase to the next, thus representing a periodic breathing cycle. These DVFs are used to warp the static volume and generate a 10-phase 4D image sequence.

\subsubsection{Training} \label{sec:training}

The model consists of just 2.2 million parameters. All models were trained using a batch size of 1, utilizing approximately 30 GB of the 48 GB available on a single NVIDIA RTX A6000. Training for 300 epochs takes approximately 100 hours. We track validation performance on a held-out portion of the dataset and select the model with the best performance. We use the Adam optimizer \cite{kingmaAdamMethodStochastic2017} with a learning rate of $10^{-5}$. The training uses MAE loss on the predicted static image and the moving 4D images with equal weights, and bending energy \cite{rueckertNonrigidRegistrationUsing1999} regularization (weight: 0.001) on the predicted DVFs. We performed basic manual hyperparameter tuning on the 30\,s synthetic dataset and used the same settings for training the 60\,s and 6\,s scan duration models.

\subsection{Evaluation}
Model performance is assessed on a synthetic test dataset and real clinical CBCT scans.

For the synthetic test dataset, where ground-truth images are available, we report Root Mean Square Error (RMSE), Peak Signal-to-Noise Ratio (PSNR), and Structural Similarity Index (SSIM) for the predicted volumes and compare them to traditional 3D (SART-TV) reconstructions. These metrics are computed on the body-only part of the image; background air and couches are removed prior to computing all metrics. Background removal is performed by thresholding the image based on HU and selecting the second-largest connected component in the image as the body (the largest component being the background).

To assess suitability for downstream clinical tasks and motion modeling performance, we use TotalSegmentator \cite{wasserthalTotalSegmentatorRobustSegmentation2023} to perform segmentation on both the ground-truth and reconstructed volumes. TotalSegmentator produces segmentation masks for over 100 structures; we focus on the relevant structures right lung, left lung, liver, heart, esophagus, and spinal cord, and report Dice similarity coefficients (referred to as Dice scores) for these segments.

Robustness and adaptability of our proposed method were evaluated by training two additional models on a 60\,s and 6\,s free-breathing setup.
Although a short 6\,s CBCT naturally exhibits fewer motion artifacts, it presents a significant challenge for motion-resolved 4D reconstruction due to the very limited number of projections per respiratory phase.

Predictions from these models were used to conduct a clinical expert evaluation. The experts were presented a side-by-side view of a SART-TV 3D reconstruction and the exhale phase image from our 4D reconstruction. They were asked to assess tumor and esophagus visibility and indicated their preferred reconstruction.

For the clinical scan test datasets, there is no ground-truth available. We provide images for qualitative assessment, comparing our method to 3D iterative algebraic (SART-TV) \cite{peterlikReducingResidualmotionArtifacts2021} and 4D McKinnon-Bates (MKB) \cite{mckinnonImagingBeatingHeart1981, lengStreakingArtifactsReduction2008} reconstructions.

\subsubsection{Input data consistency}
There is no ground-truth available for the real CBCT scans. To assess projection consistency of the predicted motion pattern with the actual motion in the scan, we assess the diaphragm motion in projection space by forward projecting our 4D reconstruction and comparing it to the original input projections. The predicted 10-phase volumes are each forward projected with acquisition settings that match the original input data, resulting in a set of 10 sinograms which are used to assess the input data consistency.

{\bf Amsterdam-Shroud method to extract diaphragm motion: }
A diaphragm displacement signal is extracted from the original input projection using the Amsterdam-Shroud (AS) \cite{zijpExtractionRespiratorySignal2004} method for the right lung. A region of interest (ROI, $13\times13\times32$ voxels) containing the diaphragm dome is manually selected for each scan. To extract the actual signal from the AS images, we use a path-finding algorithm \cite{waltScikitimageImageProcessing2014} to find the minimum-cost path through all columns of the AS image.
Due to the presence of minor residual sorting artifacts (ghost edges) from the original 4D-CT scans for the synthetic datasets (Figure \ref{fig:sim_tumor_sample}), and the slightly blurred results for our reconstruction method, we have found this to be a more robust way to extract curves from AS images than the better-known maximum-intensity or optimization-based methods. In particular, the path-finding approach also alleviates issues for scan angles where the left lung diaphragm is also visible in the AS image, which can lead to discontinuities when using other methods. Furthermore, all AS images are manually reviewed and we mark invalid regions due to the ROI leaving the field-of-view, occlusion of the diaphragm, or other artifacts.

{\bf Comparison of extracted diaphragm displacements: }
We extract AS images and diaphragm positions for our predicted max-inhale and max-exhale phase volumes. The two extracted curves represent the maximum predicted diaphragm displacement in the superior-inferior direction over all projections. To compare this to the input projections, we apply peak detection to the original AS curves and calculate the average displacement between all lower and all upper peaks. We compute the ratio of the overall displacement and a position offset for the prediction over the original input projections. The displacement ratio quantifies the predicted motion amplitude compared to the input; a value of $> 1$ means that the predicted motion is larger than the input data, while values $< 1$ mean that the motion is underestimated. The results of this method can be seen in Figures \ref{fig:sim_tumor_sample} and \ref{fig:yonsei_results}.

\subsection{Datasets}

\subsubsection{Clinical testing datasets}

We evaluated our model on three datasets acquired under different scan protocols.

\noindent \textbf{30\,s Acquisition:}
Thirteen thoracic scans from seven radiotherapy patients from Gangneung Asan Hospital, South Korea, were acquired on Varian Halcyon\texttrademark\ systems in half-fan mode (860 projections). Each scan spans 4--8 respiratory cycles. Scans were centered on the planning target volume (PTV), occasionally resulting in truncation artifacts.\\
\textit{Geometry:} source-imager distance (SID): 154\,cm, source-to-axis distance (SAD): 100\,cm, detector lateral offset: 17.5\,cm.

\noindent \textbf{60\,s Acquisition:}
Four 360\textdegree\ full-fan scans (879 projections) from individual patients, each covering 18--22 respiratory cycles.
\textit{Geometry:} SID: 154\,cm, SAD: 100\,cm.

\noindent \textbf{6\,s Acquisition:}
Five 360\textdegree\ full-fan scans (407 projections) covering 1--2 respiratory cycles.
\textit{Geometry:} SID: 154\,cm, SAD: 100\,cm.

The 60\,s and 6\,s scans were acquired using the Varian HyperSight system on a Halcyon machine with an imager size of $86\times43$\,cm (pixel size: $0.28\times1.12$\,mm).

Due to GPU memory limits, projection images were downsampled from their original size.
\textit{30\,s:} image width $366$ and height between $85$ and $147$ pixels (pixel size: $1.76 \times 2.688$\,mm);
\textit{60\,s \& 6\,s:} $768 \times 192$ pixels (pixel size: $1.12 \times 2.24$\,mm). The reconstructed volume sizes are:
\textit{30\,s:} $256 \times 256 \times 64$ voxels ($2 \times 2 \times 3$\,mm);
\textit{60\,s \& 6\,s:} $256 \times 256 \times 122$ voxels ($2.1 \times 2.1 \times 2.0$\,mm).

\subsubsection{Synthetic Datasets}
The model is trained using synthetic data generated using the method from Section \ref{sec:synthgen}.
Synthetic datasets matching each clinical setting (30\,s, 60\,s, 6\,s) were generated using a 4D-CT dataset of 78 patients (split 70\%/15\%/15\% for training/validation/testing). For each patient, 10 synthetic scans were simulated using randomly selected amplitude signals from a set of 150 traces acquired using the Varian RPM system. This resulted in 560 training, 110 validation, and 110 test samples per acquisition type.

\section{Results}
\subsection{Simulated dataset}
\subsubsection{Image quality}

\begin{figure*}[tbp]
\begin{center}
\includegraphics{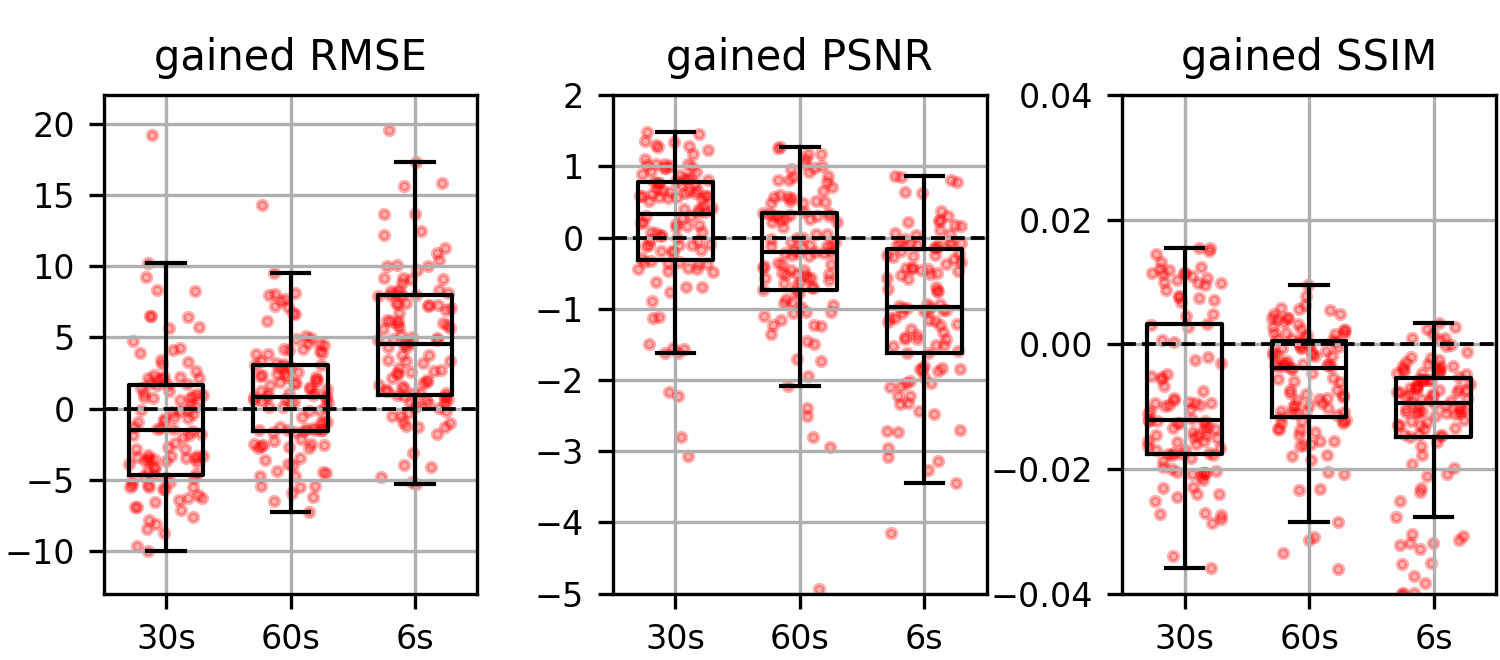}
\end{center}
\caption{\label{fig:sim_gained_imgquality} Distribution of per-sample gained image quality for the 30\,s, 60\,s and 6\,s model versions relative to SART-TV. Image quality is quantified in terms of gained RMSE (lower is better), gained PSNR (higher is better), and gained SSIM (lower is better). The boxes span the interquartile range (IQR), and the whiskers extend to the most extreme data points within $1.5\cdot\text{IQR}$. The box plots are overlaid with a scatter plot of all individual samples in the test dataset.}
\end{figure*}

\begin{figure*}[tbp]
  \centering
  \adjustbox{valign=t}{%
    \begin{minipage}[t]{0.49\textwidth}
      \includegraphics[width=\linewidth]{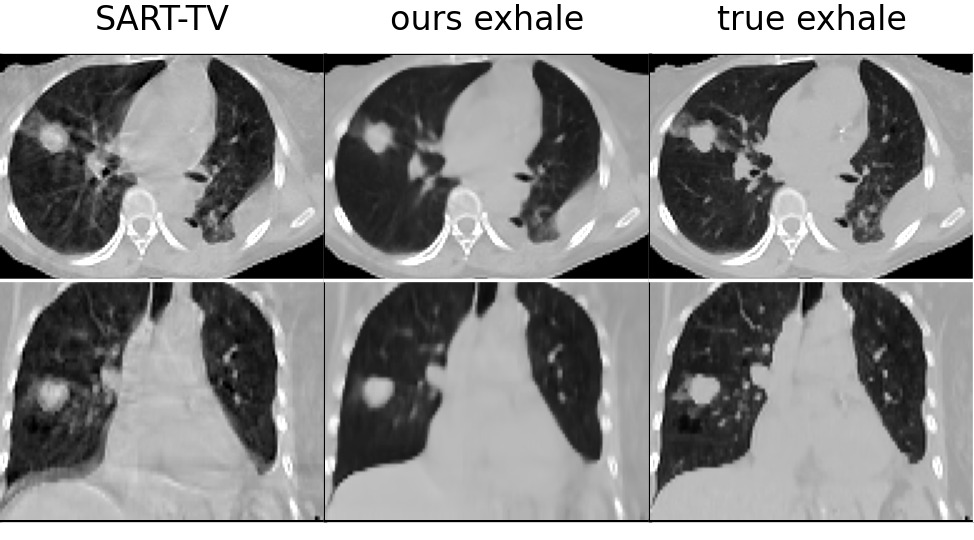}\\[1mm]
      \includegraphics[width=\linewidth]{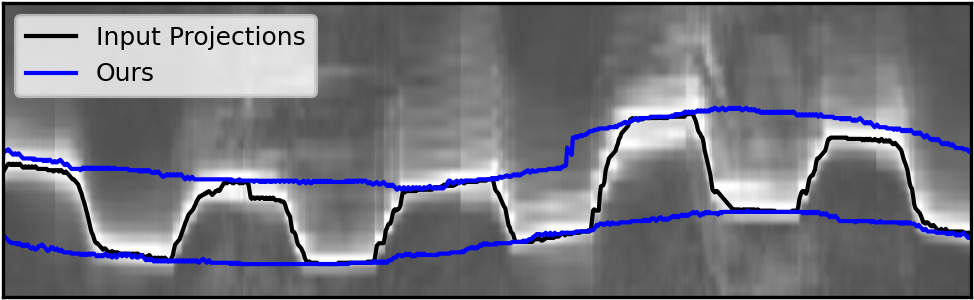}
    \end{minipage}%
  }%
  \hspace{3mm}%
  \adjustbox{valign=t}{%
    \begin{minipage}[t]{0.49\textwidth}
      \includegraphics[width=\linewidth]{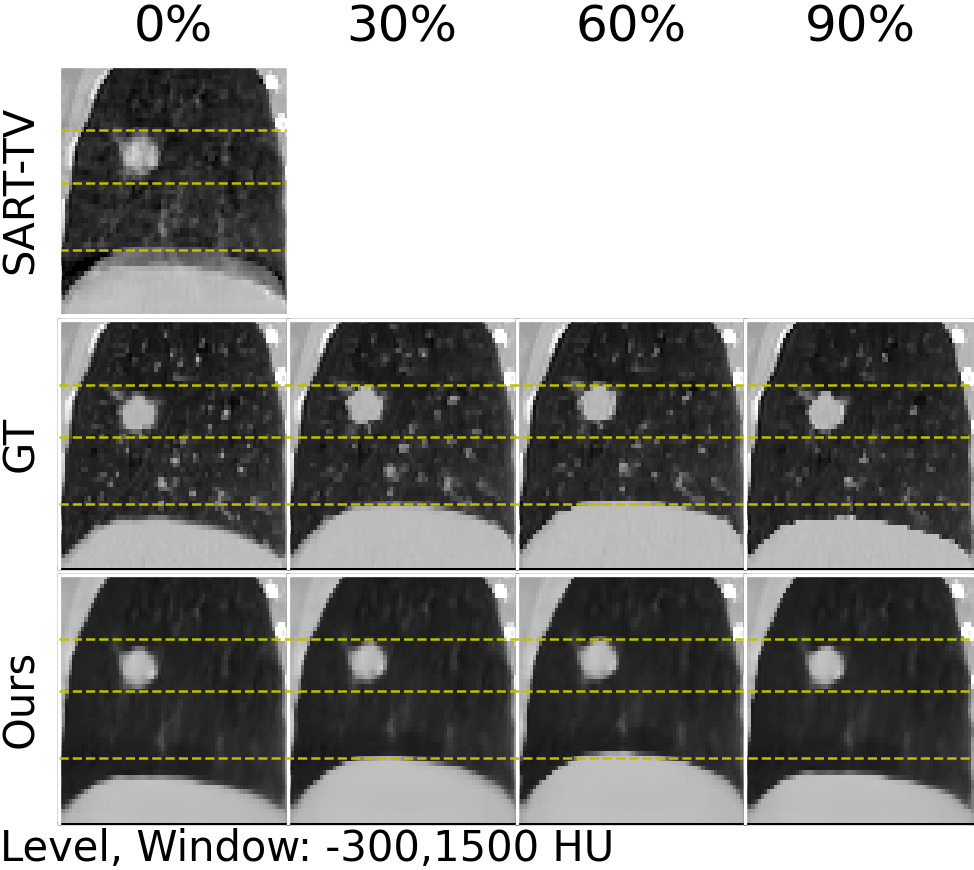}
    \end{minipage}%
  }%
\caption{Sample from synthetic test dataset with visible lung tumor.
\emph{Top left}: Image quality comparison of our method with SART-TV and ground-truth image, showing reduced motion artifacts.
\emph{Right}: 4D reconstruction of our method compared to the ground-truth, and a 3D SART-TV reconstruction. Shown are four out of 10 respiratory phases.
\emph{Bottom left}: Extracted diaphragm motion (AS) from the original input projections and in-/exhale phases from our 4D reconstruction. The reconstructed phases are consistent with the in-/exhale positions.}
\label{fig:sim_tumor_sample}
\end{figure*}

Table \ref{tab:simresults} presents the mean RMSE, PSNR, and SSIM performance metrics for our 4D reconstruction method compared to the SART-TV 3D reconstruction on the simulated test datasets. To ensure a conservative comparison, SART-TV 3D metrics were computed against all ten ground-truth phases, and only the best-matching (maximum) value was used. Our 4D results are reported as the average value over all ten phases.

The image quality metrics depend on intrinsic scan properties, such as motion amplitude and tissue distribution. Figure \ref{fig:sim_gained_imgquality} shows the distribution of per-sample \emph{gained} image quality of our method compared to SART-TV. The 60\,s and 6\,s retrained models exhibit a slight quantitative decrease, likely because initial model development and hyperparameter tuning were focused on the 30\,s protocol (see Section \ref{sec:training}). Furthermore, the 6\,s scan presents an extra challenge, providing only 1--2 full respiratory periods for motion inference. Overall, these results show that our method can achieve the benefits of 4D motion resolution with quantitative image quality close to a full-projection-count 3D reconstruction.

As illustrated in Figure \ref{fig:sim_tumor_sample}, motion streak artifacts present in the SART-TV reconstruction are reduced, and tissue interfaces of moving structures (diaphragm, tumor) are better defined. However, perceived sharpness and visibility of fine details and small structures are reduced, giving the predicted image a blurred appearance. This blurring effect is a known limitation of networks trained with pixel-wise loss functions such as MAE \cite{ghodratiMREvaluationLossFnc}.

The predicted tumor and diaphragm motion over the breathing cycle matches the positions in the ground-truth images. In the AS image, the diaphragm position of the predicted in-/exhale phases matches the peak positions of the original input projections. However, compared to AS images of real CBCT scans, the diaphragm trace curve is flatter, and there are some smeared-out or ghost edges visible in the image. These effects stem from the synthetic CBCT and motion simulation framework used in this study.

\begin{table}[hb]
\centering
\setlength{\tabcolsep}{4pt}
\renewcommand{\arraystretch}{1.3}
\begin{tabular}{|@{}l@{}|r|r|r|}
\hline
 & \textbf{SART-TV} & \textbf{Ours(Static)} & \textbf{Ours(4D)} \\
\hline
\textbf{30\,s} & & & \\

RMSE & 47.64$\pm$9.98  & 49.67$\pm$8.39  & 46.45$\pm$7.13 \\
PSNR & 32.70$\pm$2.16  & 32.22$\pm$1.46 & 32.79$\pm$1.37 \\
SSIM & 0.956$\pm$0.017 & 0.943$\pm$0.015 & 0.947$\pm$0.014 \\

\textbf{60\,s} & & & \\

RMSE & 38.29$\pm$7.28  & 61.95$\pm$7.13  &39.16$\pm$5.41 \\
PSNR & 34.53$\pm$1.77  & 30.24$\pm$1.00  &34.25$\pm$1.21 \\
SSIM & 0.961$\pm$0.015 & 0.935$\pm$0.016 &0.954$\pm$0.014 \\

\textbf{6\,s} & & & \\

RMSE & 39.79$\pm$7.93  & 84.59$\pm$6.88  & 44.39$\pm$7.86 \\
PSNR & 34.20$\pm$1.77  & 27.50$\pm$0.72  & 33.21$\pm$1.57 \\
SSIM & 0.967$\pm$0.012 & 0.911$\pm$0.018 & 0.955$\pm$0.015 \\
\hline
\end{tabular}
\caption{\label{tab:simresults} Image quality metrics for the simulated dataset (RMSE in HU, PSNR in dB).}
\end{table}

\subsubsection{Segmentation Results}
\begin{figure*}[tbp]
\begin{center}
\includegraphics{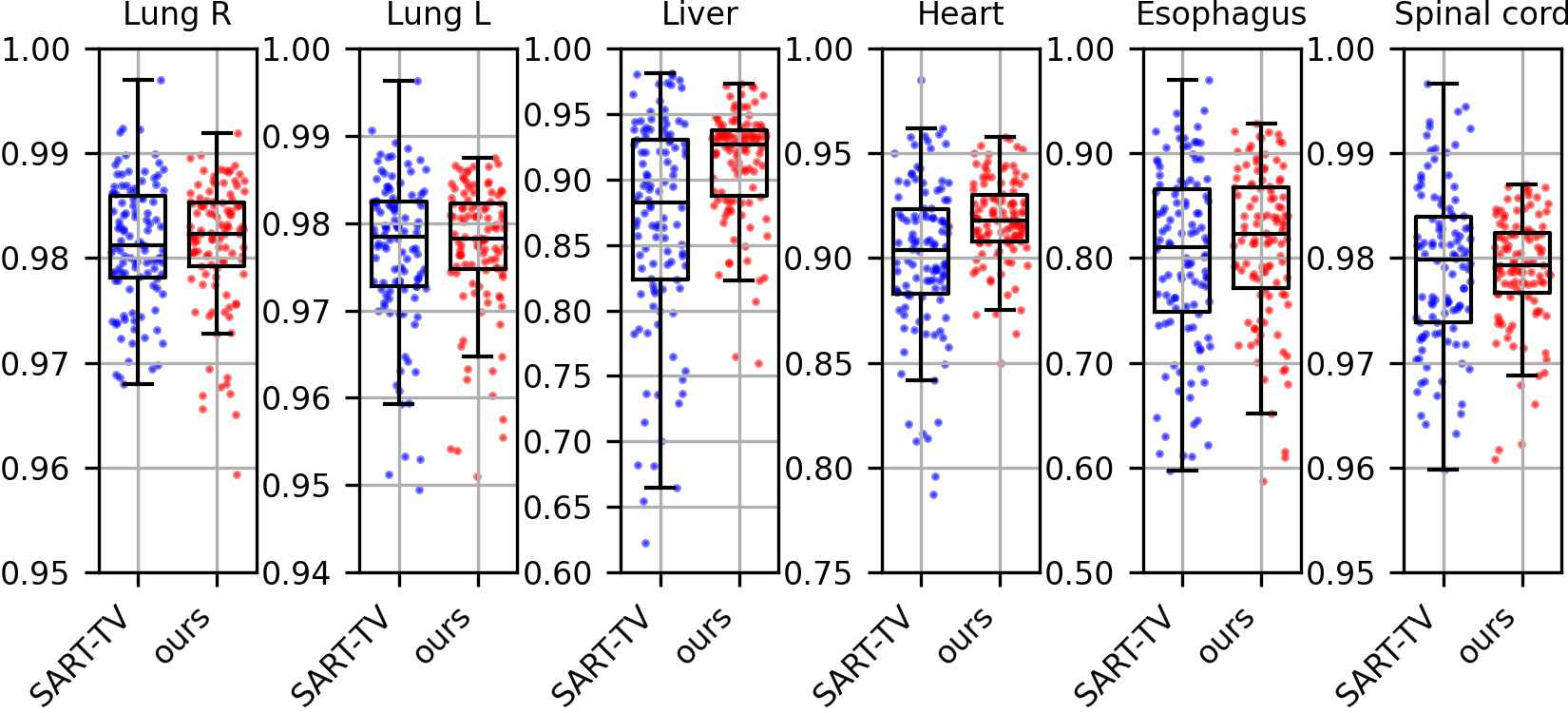}
\end{center}
\caption{\label{fig:sim_dice_scores} Segmentation Dice scores for SART-TV and our method, showing improved segmentation performance for segments affected by respiratory motion (lungs, liver, heart, esophagus) and consistent segmentation scores for the stationary spinal cord.}
\end{figure*}

Mean segmentation Dice scores for segments most affected by respiratory motion are consistently higher compared to SART-TV (right lung: +0.0089, left lung: +0.0077, liver: +0.10, heart: +0.028, esophagus: +0.032) and stable for the stationary spinal cord (+0.0032). Figure \ref{fig:sim_dice_scores} shows the distribution of Dice scores for the samples in the test dataset. The Dice scores for SART-TV are the maximum scores against all 10 ground-truth phases, while for our 4D method Dice scores are averaged over all 10~phases.

\subsection{Clinical scan dataset}

\begin{figure}
\centering
\includegraphics{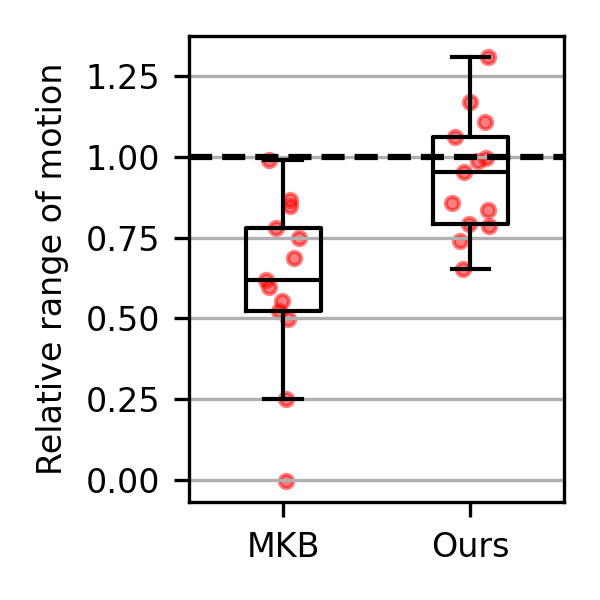}
\caption{\label{fig:yonsei_rel_displacement} Relative diaphragm displacement for the 30\,s clinical test dataset for the MKB reconstruction and our proposed CNN method. A value of 1.0 means that the predicted diaphragm displacement exactly matches the displacement in the original X-ray projections; values $> 1$ or $< 1$ mean the displacement is over- or underestimated, respectively.}
\end{figure}

\begin{figure*}[tbp]
  \centering
  \adjustbox{valign=t}{%
    \begin{minipage}[t]{0.49\textwidth}
      \includegraphics[width=\linewidth]{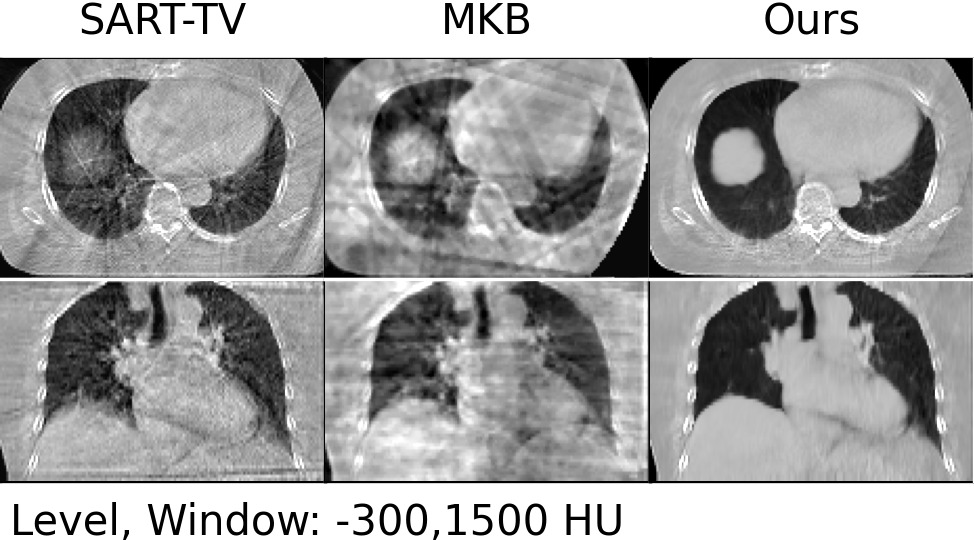}
    \end{minipage}%
  }%
  \hspace{3mm}%
  \adjustbox{valign=t}{%
    \begin{minipage}[t]{0.49\textwidth}
      \includegraphics[width=\linewidth]{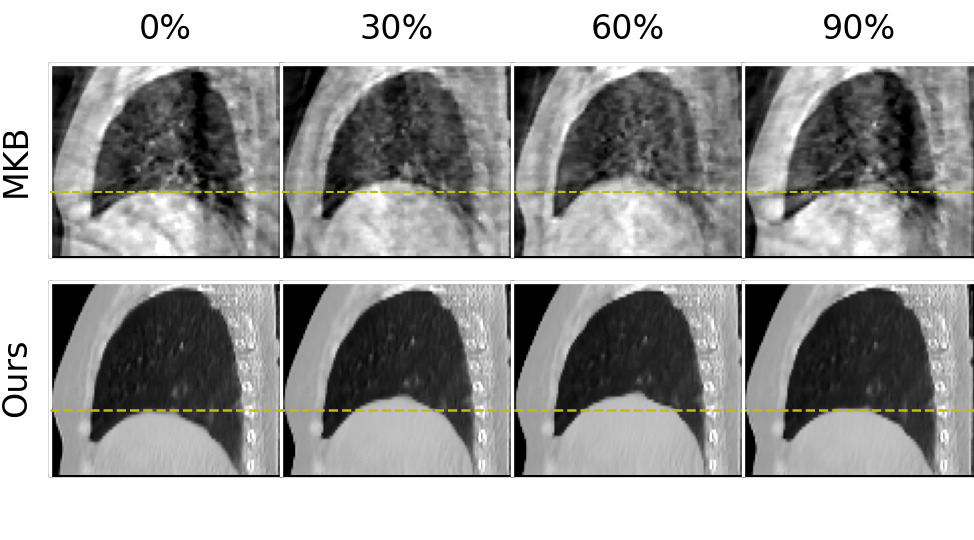}
    \end{minipage}%
  }
  \includegraphics{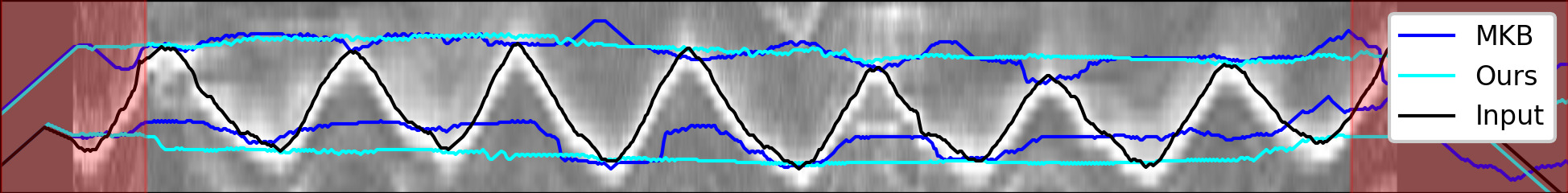}
\caption{Sample from clinical test dataset.
\emph{Top left}: Image quality comparison of our method with 3D SART-TV and 4D MKB reconstructions. Motion streaking and sparse sampling artifacts are reduced by our method.
\emph{Top right}: 4D reconstruction of our method compared to MKB. Shown are four out of ten respiratory phases.
\emph{Bottom}: Extracted diaphragm motion of the input projections and from the in-/exhale phases from our 4D and a 4D MKB reconstruction. The reconstructed phases match the in-/exhale position of the original input projections for both MKB and our methods.}
\label{fig:yonsei_results}
\end{figure*}

Ground-truth is not available for the real clinical scans. Figure \ref{fig:yonsei_results} provides a visual comparison of reconstructed image quality and motion fidelity across respiratory phases. Our method shows improved overall image quality compared to SART-TV and MKB reconstructions. Motion streak artifacts are greatly reduced and moving structures are better defined than in SART-TV reconstructions. In addition, there are no visible sparse sampling artifacts compared to MKB.

Compared to MKB, our method preserves respiratory motion while reducing noise and sparse sampling artifacts. Diaphragm tracking results show that predicted motion is consistent with the input projections (Figure \ref{fig:yonsei_results}).
Quantitatively, in terms of predicted diaphragm displacement compared to the input projections, our method predicts diaphragm motion better than MKB, which often underestimates the motion (Figure \ref{fig:yonsei_rel_displacement}). Relative diaphragm displacement: MKB mean 0.58 (std 0.1), ours mean 0.98 (std 0.1).

\subsection{Retrained models for different acquisition geometries}

Our trained models are inherently sensitive to CBCT acquisition parameters such as scan duration and projection count, which affect the spatial and temporal resolution of the acquired data.
To evaluate robustness across different CBCT protocols, we retrained our model using data matched to the geometry of HyperSight 60\,s and 6\,s acquisitions, maintaining the same network architecture and training configuration as used for the 30\,s model.

The 30\,s model achieved the best image quality in terms of RMSE and PSNR. The 60\,s and 6\,s models performed slightly worse (mean PSNR differences of $-0.28$\,dB and $-0.99$\,dB, respectively, compared to SART-TV). This may reflect that model development and hyperparameter tuning were performed for the 30\,s setting only. No protocol-specific adjustments were made for the 60\,s and 6\,s models, indicating potential for improvement through targeted optimization.

For the 60\,s and 6\,s scans, we observe a larger difference in RMSE and PSNR between the predicted static image and the final (warped) 4D reconstruction than for the 30\,s scan (60\,s: +4.01\,dB; 6\,s: +5.71\,dB; 30\,s: +0.57\,dB), suggesting that the networks not only learn to use the predicted DVFs to represent actual patient motion but also use the warping step to increase image quality. This further suggests that there is room for improvement through hyperparameter tuning and DVF regularization.

\begin{figure*}[tbp]
  \centering
  \adjustbox{valign=t}{%
    \begin{minipage}[t]{0.49\textwidth}
      \includegraphics[width=\linewidth]{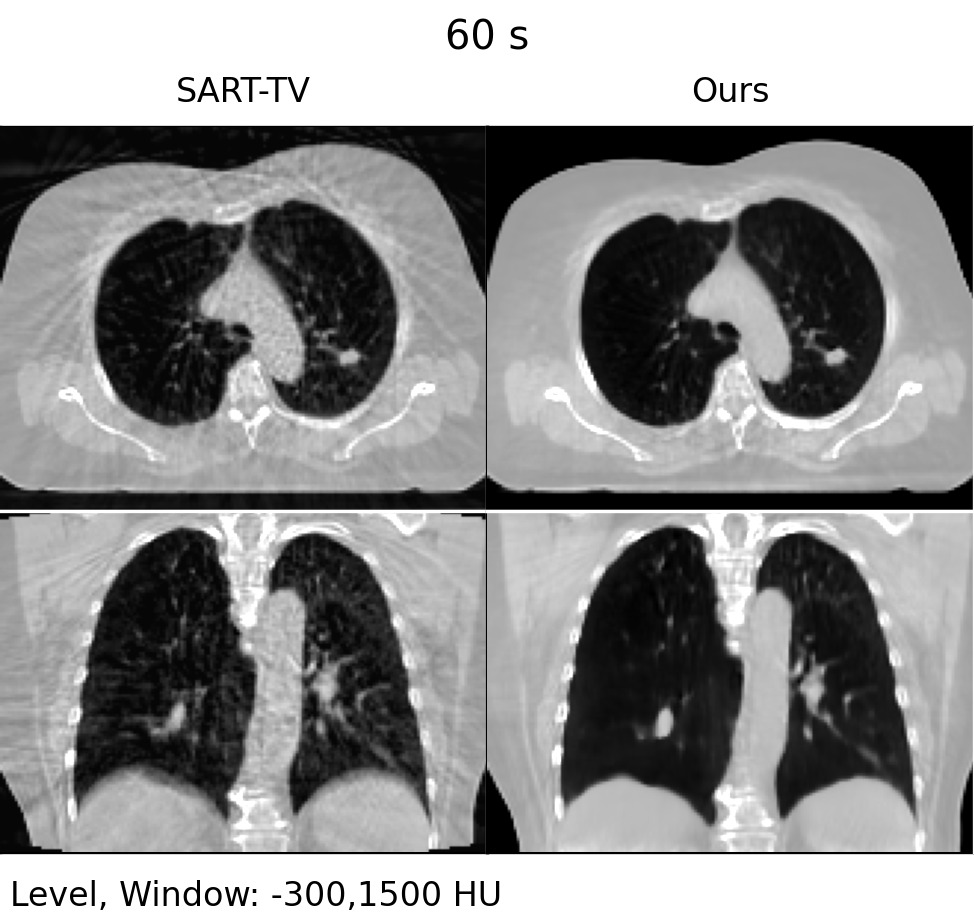}
      \includegraphics[width=\linewidth]{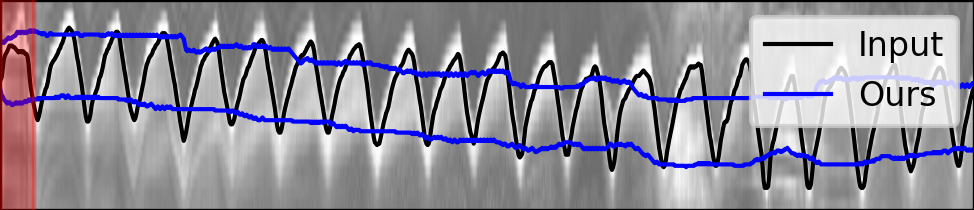}
    \end{minipage}%
  }%
  \hspace{3mm}%
  \adjustbox{valign=t}{%
    \begin{minipage}[t]{0.49\textwidth}
      \includegraphics[width=\linewidth]{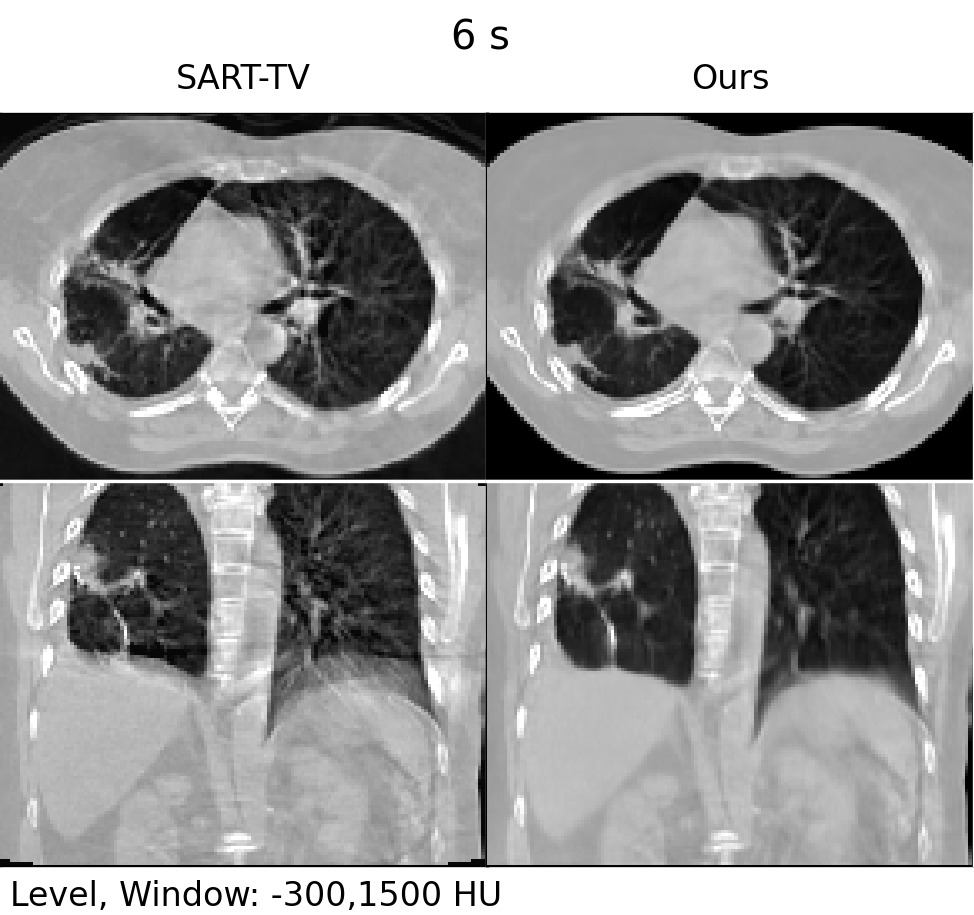}
      \includegraphics[width=\linewidth]{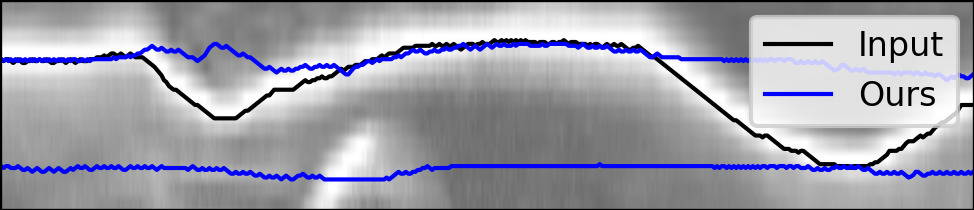}
    \end{minipage}%
  }
\caption{Comparison of SART-TV 3D to our 4D reconstruction for two HyperSight scans with AS images and diaphragm positions (left: 60\,s, right: 6\,s).}
\label{fig:hypersight_results}
\end{figure*}

Despite the slight decrease in image quality metrics, the model still produced 4D volumes with reduced motion artifacts and improved anatomical boundary definition. The extracted diaphragm motion traces showed good agreement with the input projection data, indicating robust motion estimation (Figure \ref{fig:hypersight_results}).

\begin{figure*}[tb]
\centering
\makebox[\textwidth][c]{%
\begin{minipage}{0.44\textwidth}
\includegraphics[width=\linewidth]{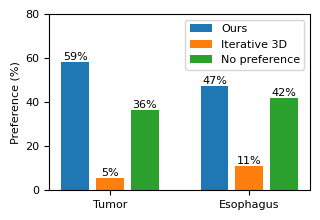}
\captionof{figure}{\label{fig:clinical_eval_bars} Clinical expert evaluation for the two tasks: tumor visibility and esophagus visibility. The bars show the indicated preference of the clinical experts for our 4D or a SART-TV 3D reconstruction method, averaged over all five scans.}
\end{minipage}%
\hspace{2mm}
\begin{minipage}{0.45\textwidth}
\small
\renewcommand{\arraystretch}{1.5}
\begin{tabular}{|@{}p{0.65cm}@{}|@{}p{0.91cm}@{}@{}p{1.2cm}@{}@{}p{0.8cm}@{}|@{}p{0.98cm}@{}@{}p{1.3cm}@{}@{}c|}
\hline
 & \multicolumn{3}{c|}{\textbf{Tumor visibility}} & \multicolumn{3}{c|}{\textbf{Esophagus visibility}} \\
&Ours&Iter.\,3D&No pref.&Ours&Iter.\,3D&No pref.\\ \hline
6\,s & \centering 10 & \centering 1 & \centering 0 & \centering  4 & \centering  1 & 6 \\ \hline
6\,s & \centering 4 & \centering 0 & \centering 7 & \centering  5 & \centering  1 & 5 \\ \hline
60\,s & \centering 7 & \centering 0 & \centering 4 & \centering  8 & \centering  1 & 2 \\ \hline
60\,s & \centering 5 & \centering 2 & \centering 4 & \centering  4 & \centering  1 & 6 \\ \hline
60\,s & \centering 6 & \centering 0 & \centering 5 & \centering  5 & \centering  2 & 4 \\ \hline
\end{tabular}
\captionof{table}{\label{tab:clinicaleval} Results for all five cases from the clinical evaluation. The table lists two 6\,s and three 60\,s scans and the number of experts (out of a total of 11) who preferred our 4D reconstruction, an iterative 3D reconstruction, or indicated no preference.}
\end{minipage}
}
\end{figure*}

\begin{figure*}[t]
\includegraphics{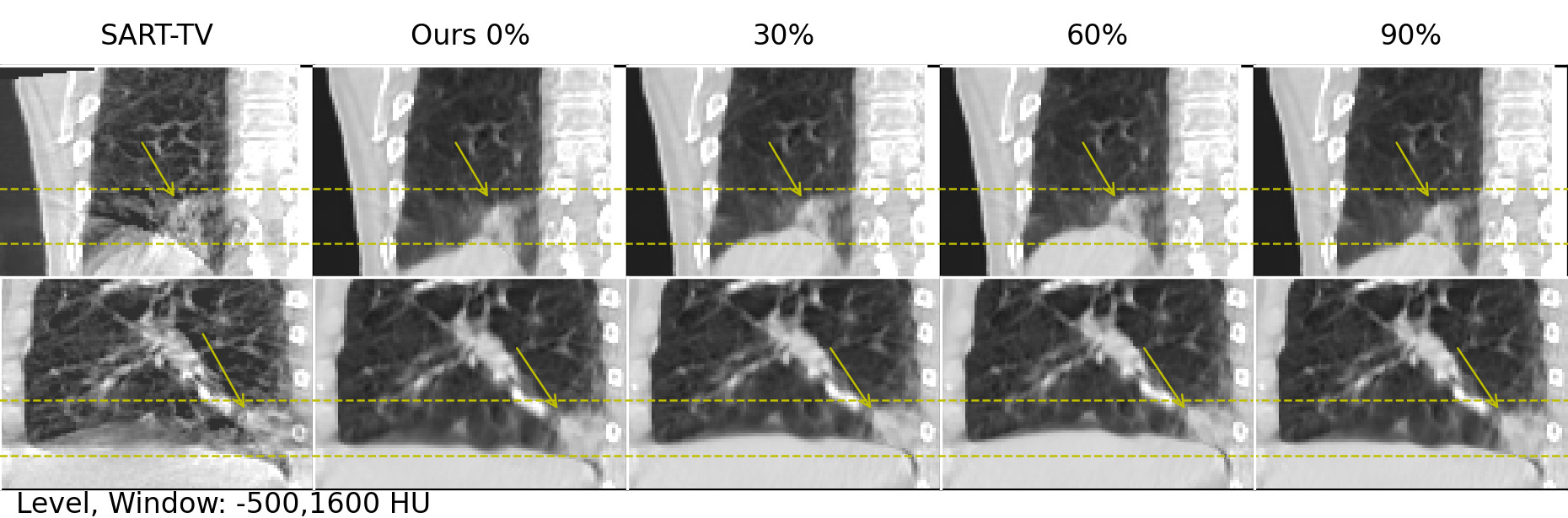}
\caption{\label{fig:6s_tumor} Comparison of SART-TV 3D to our 4D reconstruction of a HyperSight 6\,s scan of a case with a tumor in the lower right lung. Improved tumor visibility and delineation of our 4D vs.\ 3D SART-TV reconstruction.
This case was presented in the expert evaluation (first row of Table \ref{tab:clinicaleval}), with 10 out of 11 experts agreeing on improved tumor visibility for our 4D reconstruction.}
\end{figure*}

\subsection{Clinical Expert Evaluation}

A total of eleven clinical experts (seven radiation oncologists and four medical physicists) participated in the evaluation. Each expert assessed two images from the HyperSight 6\,s dataset and three images from the HyperSight 60\,s dataset. The experts were shown side-by-side a SART-TV 3D image and a single static image (exhale phase) of our 4D reconstruction. They were asked to select their preferred image for (a) tumor visibility and (b) esophagus visibility.

For tumor visibility, averaged over all five samples and experts, 59\% of votes indicated a preference for our reconstruction, 36\% indicated no preference, and 5\% favored the iterative 3D reconstruction. Similarly, for esophagus visibility, 47\% preferred our reconstruction, while 42\% had no preference and 11\% favored the iterative 3D approach (Figure \ref{fig:clinical_eval_bars} and Table \ref{tab:clinicaleval}). The high percentage of "no preference" votes suggests that our 4D reconstruction provides comparable structural visibility to the 3D images, while also adding temporal information. 
Figure \ref{fig:6s_tumor} shows a scan from a difficult case with a lung tumor in the right lower lung. In the SART-TV 3D reconstruction, tumor visibility is hindered by strong streaking artifacts. In comparison, our proposed 4D method reduces the streaking artifacts and provides a motion-resolved reconstruction, with 10 out of 11 experts agreeing on better tumor visibility.

\FloatBarrier
\section{Discussion}
We proposed a CNN-based 4D CBCT reconstruction method and evaluated its performance across simulated and clinical datasets. On simulated data, our method achieved image quality metrics comparable to conventional 3D reconstruction techniques, while additionally providing motion resolution across respiratory phases. However, our reconstruction results exhibit a blurring effect that can obscure high-frequency details. This is a known limitation of training with pixel-wise loss functions such as Mean Absolute Error (MAE). To improve sharpness, future investigations could explore alternative or combined loss formulations, such as SSIM-based or perceptual losses, to improve image sharpness while preserving robustness to motion artifacts.

In clinical scenarios, where ground-truth is unavailable, we assessed the method through qualitative visual comparison and diaphragm motion analysis using AS images. Our approach substantially reduced sparse sampling artifacts and increased motion fidelity compared to MKB reconstructions. A key advantage of our method is that it does not require external respiratory signals, signal extraction, or any explicit binning of projections. Instead, the network infers temporal phase information directly from the projection data.

We demonstrated adaptability to different acquisition protocols by retraining the model to match HyperSight 6\,s and 60\,s scans, suggesting potential for integration into different clinical workflows. Even under high-speed, sparse-view conditions, the method produced motion-resolved volumes that aligned well with expected motion patterns, as indicated by diaphragm trajectory estimates. The results for HyperSight 6\,s scans are especially relevant, since current motion modeling or 4D reconstruction approaches generally require longer scans covering multiple respiratory periods, in contrast to just 1--2 full periods in the 6\,s scans. Traditional 4D CBCT requires scan times of 60 or more seconds. Our study shows 4D reconstructions from 60\,s, 30\,s, and even 6\,s scans (a reduction of 50\% to 90\%). Our method does not require a recorded surrogate signal, which can further reduce the required setup time.
Furthermore, standard 4D acquisition protocols often utilize higher mAs settings to compensate for the image quality degradation caused by projection binning. For instance, a Varian Halcyon standard 4D thorax protocol delivers 11.1\,mGy compared to 4.45\,mGy for a 3D scan ($CTDI_{vol}$, 32\,cm), a dose reduction of roughly 60\%.

A clinical expert evaluation reinforced our quantitative findings. Across five patient cases, the majority of experts preferred our reconstructions over SART-TV in terms of both tumor and esophagus visibility. This consensus was particularly pronounced in the challenging case of a lower-lung tumor, where strong streaking artifacts in the SART-TV reconstruction obscured the target, while our method yielded much-improved tumor visibility. We recognize that our clinical evaluation is limited by the number of experts and by using only five patient cases; it should therefore be seen as a preliminary validation of our method. Nevertheless, the results of the study suggest that our method successfully enables time-resolved 4D reconstructions without impacting perceived image quality compared to traditional 3D reconstructions.

Direct validation of the predicted motion remains challenging due to the lack of ground-truth. Our approach of comparing the motion present in the original input projections to the predicted diaphragm motion states in the AS images provides a useful proxy to judge motion fidelity. However, additional studies may be necessary for more rigorous motion validation. Our findings reveal that quantitative image quality metrics are consistently lower for the single predicted static volume but improve substantially when DVFs are applied. This observation suggests that the model not only leverages the DVFs for motion modeling but also implicitly enhances image quality through their application.

Image quality assessment is relatively straightforward through established quantitative metrics, whereas motion modeling lacks equally accessible and reliable evaluation criteria. This potential trade-off complicates efforts to fine-tune model hyperparameters.

\section{Conclusion}
We introduced a novel CNN-based U-Net architecture for 4D CBCT reconstruction directly from projection data, eliminating the need for recorded respiratory signals or explicit projection binning.

The model is trained on a population dataset, requiring no patient-specific adaptation. We implemented this as a compact network with just over 2 million parameters, enabling 10-phase 4D reconstruction on a single mid-tier GPU in under 4 seconds. This efficient approach offers high-quality, motion-resolved imaging and demonstrates strong potential for integration into clinical workflows by reducing complexity, setup time, and reliance on respiratory surrogates.

\section*{Acknowledgments}
\small
\noindent Clinical test dataset (30\,s scans) supplied by Yonsei University and Gangneung Asan Hospital, University of Ulsan College of Medicine, IRB approval number GNAH IRB 2022-07-008.

\noindent HyperSight 60\,s and 6\,s datasets supplied by Maastro Clinic and Nova Scotia Health.

\noindent We also thank Yonsei University for facilitating the clinical expert evaluation.

\noindent This work was supported by Innosuisse under grant 120.480 IP-LS.

\section*{Conflict of Interest Statement}
\small
Pascal Paysan, Igor Peterlik, and Michal Walczak are employees of Varian Medical Systems Imaging Laboratory GmbH.

\bibliography{bib}
\bibliographystyle{medphy}
\end{document}